\documentclass[review]{elsarticle}

\usepackage{lineno,hyperref}
\usepackage{graphicx,subfigure}
\usepackage{multirow}
\usepackage{xcolor}

\begin{document}
	
	\begin{frontmatter}
		
		\title{Development of a Dataset and a Deep Learning Baseline Named Entity Recognizer for Three Low Resource Languages: Bhojpuri, Maithili and Magahi}

		\author[1]{Rajesh Kumar Mundotiya}
		\ead{rajeshkm.rs.cse16@iitbhu.ac.in}
		
		\author[2]{Shantanu Kumar}
		\ead{shantanubhu97@gmail.com}
		
		\author[3]{Ajeet kumar}
		\ead{ajeet786sanjay@gmail.com}
		
		\author[3]{Umesh Chandra Chaudhary}
		\ead{umeshchandra8294@gmail.com}
		
		\author[2]{Supriya Chauhan}
		\ead{supriyachauhanaisha@gmail.com}
		
		\author[4]{Swasti Mishra}
		\ead{swasti.hss@iitbhu.ac.in}
		
		\author[2]{Praveen Gatla}
		\ead{praveengatla@gmail.com}
		
		\author[1]{Anil Kumar Singh}
		\ead{aksingh.cse@iitbhu.ac.in}
		
		\address[1]{Department of Computer Science and Engineering, IIT(BHU), India}
		\address[2]{Department of Linguistics, Banaras Hindu University, India}
		\address[3]{Cognizant, Bangalore, India}
		\address[4]{Department of Humanistic Studies, IIT(BHU), India}
		
		
		\begin{abstract}
			In Natural Language Processing (NLP) pipelines, Named Entity Recognition (NER) is one of the preliminary problems, which marks proper nouns and other named entities such as Location, Person, Organization, Disease etc. Such entities, without a NER module, adversely affect the performance of a machine translation system. NER helps in overcoming this problem by recognising and handling such entities separately, although it can be useful in Information Extraction systems also. Bhojpuri, Maithili and Magahi are low resource languages, usually known as Purvanchal languages. This paper focuses on the development of a NER benchmark dataset for the Machine Translation systems developed to translate from these languages to Hindi by annotating parts of their available corpora. Bhojpuri, Maithili and Magahi corpora of sizes 228373, 157468 and 56190 tokens, respectively, were annotated using 22 entity labels. The annotation considers coarse-grained annotation labels followed by the tagset used in one of the Hindi NER datasets. We also report a Deep Learning based baseline that uses an LSTM-CNNs-CRF model. The lower baseline F$_1$-scores from the NER tool obtained by using Conditional Random Fields models are 96.73 for Bhojpuri, 93.33 for Maithili and 95.04 for Magahi. The Deep Learning-based technique (LSTM-CNNs-CRF) achieved 96.25 for Bhojpuri, 93.33 for Maithili and 95.44 for Magahi.
		\end{abstract}
		
		
	\end{frontmatter}
	

	\section{Introduction}
	\label{intro}
	Named Entity Recognition (NER) is the process of identification of named entities (Person, Organization, Location etc) in natural language text. The present paper concentrates on three low resource languages (LRLs): Bhojpuri, Maithili and Magahi (BMM), which belong to the Indo-Aryan language family. This work may be seen as the first attempt to develop an NER tool for Bhojpuri, Maithili and Magahi. There is no previous work on NER for these languages as far as we know. The main aim of the present paper is to start with insights from the NER systems that are developed for Indian Languages with more resources and based on that we try to develop an NER System for BMM.
	
	The NER module can be an important component  in  Natural  Language  Processing and Information Extraction systems.  It is an essential task for computational purposes like Machine Translation (MT), developing search engines, automatic indexing, document classification  and  text  summarization, questiona answering etc., because it is not possible to build end-to-end Deep Learning systems for these languages due to the lack of data. It  will also  be helpful  in  many  cross-linguistic  applications  as  it is relevant for other Indian Languages, particularly LRLs. The present study mainly focuses on Named Entities (NE) for BMM with machine translation as the goal.
	
	\subsection{Named Entity Recognition} 
	The concept of Named Entity was introduced in the Sixth Message of Understanding Conference (MUC-6)~\cite{grishman1996message}. It was often seen as part of an Information Extraction system, which refers to the automatic extraction of structured information such as entities, relationships between entities and attributes describing entities from unstructured sources. The role of NER system is to locate and classify words in a text into predefined categories such as the names of persons, organizations, locations, expressions of times, quantities etc. The NEs could be identified in two conventional ways, before the recent success of machine learning and then Deep Learning based techniques:
	
	\begin{enumerate}
		\item A  raw sentence was compared with gazetteer’s lists to identify the NEs, where the gazetteer’s lists are created manually for person names, location names and organization names etc.
		\item One may identify the Named Entities based on language specific linguistic rules. For example, proper nouns always start with capital letters in English, e.g. London, Shakespeare, Darwin etc.
	\end{enumerate}
	
	It is a challenging task to implement NER for Indian languages due to the absence of capitalization in their writing systems. On the other hand, these systems are phonetically organized and designed, which makes it easily possible to use phonetic features for NER for Indian languages. Preparing a gazetteer’s list for all nouns is impossible because there can be a vast number of unknown named entities in the world in terms of a corpus versus a language. Here, one important point to be noted is that not much work has been reported for NER for Low Resource languages due to insufficient lexical resources and also due to morphological richness. There have been efforts on major Indian languages, i.e., Hindi, Tamil, Telugu, Urdu, Punjabi, but no efforts on Low Resource Indian languages such as BMM.
	
	\subsection{Bhojpuri, Maithili, Magahi : An Introduction}
	
	Bhojpuri is often considered a major `sub-language' of Hindi. It is not only a language which is spoken in various states of India but in other countries as well, viz. Nepal, Mauritius, Fiji, Surinam etc. The writing system of Bhojpuri was earlier Kaithi script but now Devanagari script is used more to write Bhojpuri. According to 2011 census~\footnote{\url{https://www.censusindia.gov.in/2011Census/Language-2011/Statement-1.pdf}}, there are 5,05,79,447 Bhojpuri speakers.
	
	Maithili belongs to the Indo-Aryan language family, while Bhojpuri and Magahi are considered `sub-languages' (or even dialects) of Hindi and are mainly spoken in Eastern Uttar Pradesh, Bihar and Jharkhand states of India. Maithili is included in the 22 `scheduled' languages of the Republic of India (1950, Constitution, Article 343). Maithili was added in the Constitution of India in 2003 by the 92nd Constitutional Amendment Act. Maithili, a sister language of Hindi, is spoken in India, particularly in Bihar, Jharkhand, Uttar Pradesh etc. as well as in Nepal. It is the only language in the Bihari sub-family that is included in the eighth schedule of the Indian constitution. There are 1,35,83,464 Maithili speakers (Census, 2011). It is also one of the 122 recognised languages of Nepal. In 2007, Maithili was included in the interim Constitution of Nepal and in March 2018, it received the second official language status in the Jharkhand state of India. It too was earlier considered a sub-language or a dialect.
	
	Magahi or Magadhi, also considered a major sub-language of Hindi, is chiefly spoken in some districts of Bihar, Jharkhand, and also in the Maldah district of West Bengal. Magahi was also written in the Kaithi script in earlier days, but at present it is usually written in the Devanagari script. There are 1,27,06,825 Magahi speakers (Census, 2011).
	
	Earlier work on machine translation (particularly rule-based or transfer-based) has reported that proper handling of named tokens can improve the translation quality and performance~\cite{babych2003improving,bhalla2013improving,vu2020korean}. These named tokens would have been (mis)translated during source to target translation without an NER module, but with an NER module they can instead be simply transliterated. The current BMM machine translation systems for which we plan to use our NER module, is based on a transfer-based approach to machine translation. Even though the MT systems are based on a transfer approach, the NER module (like the POS tagging and Chunking modules) can be based on machine learning or Deep Learning, not a rule-based approach. Due to this, we have annotated some corpus and developed an NER system for these three languages and have reported the lower and a higher baseline results. The former is based on CRF and the latter on a combination of Long Short Term Memory (LSTM), Convolutional Neural Networ (CNN) and Conditional Randon Fields (CRF), called LSTM-CNNs-CRF.
	
	\subsection{Contributions}
	As there is no prior work on the NER problem for Bhojpuri, Maithili and Magahi, the contributions in this paper are as follows:
	\begin{itemize}
		\item Annotation of NEs in Bhojpuri, Maithili and Magahi corpora, with the sizes being 228373, 157468 and 56190 tokens, respectively, using 22 entity labels at the fine-grained level.
		\item Provide benchmarking results (F$_1$-score) on these annotated datasets by using a conventional machine learning technique (CRF).
		\item Apply a state-of-the-art technique (LSTM-CNNs-CRF) to improve the benchmarking scores and provide an upper baseline for future NER for these languages.
	\end{itemize}

	\section{Related Work}
	\label{related}
	
	An NER module can be a part of several Natural Language Processing and Understanding systems. Earlier work relied on rule-based techniques, which used orthographic features, lexicons and ontologies. Rau et al.~\cite{rau1991extracting} reported an initial work for extracting company names by using heuristic and hand-crafted feature-based algorithm. Later, feature engineering techniques evolved with machine learning. Weak supervision was also a promising approach, so a bootstrapping method was used, which found contextual patterns through seed entities and ranked them ~\cite{riloff1999learning}. However, more accurate contextual information could be gathered from syntactic relations ~\cite{cucchiarelli2001unsupervised}. Pasca et al.~\cite{pasca2006organizing} generated synonyms by distributional similarity for generalized contextual patterns. Zhang et al.~\cite{zhang2013unsupervised} used Inverse Document Frequency (IDF) and shallow syntactic knowledge, which filters lower IDF ranked terms before the prediction of the classifier. Apart from this, pointwise mutual information that is commonly used for Information Retrieval, was exploited to classify name entities~\cite{etzioni2005unsupervised}.
	
	WordNet~\cite{miller1995wordnet} was also employed for labelling NEs by selecting the NE synsets from WordNet, based on the frequent appearance of entities in the corpus ~\cite{alfonseca2002unsupervised}.
	
	The utility of machine learning techniques was seen in the CONLL-2003 shared task~\footnote{https://www.clips.uantwerpen.be/conll2003/ner/}, organised on four different languages: Spanish, Dutch, German and English, each with four entities (Person, Location, Organization and Miscellaneous). On the CONLL task, various machine learning techniques have been evaluated which cover AdaBoost, Hidden Markov Model, Maximum Entropy, Conditional Random Fields (CRF), Memory-based Learning, Transformation-based learning, Support Vector Machine (SVM), recurrent neural networks, Voted Perceptron and combinations of them with rules or handcrafted features~\footnote{http://www.cnts.ua.ac.be/conll2003/ner/}. Similarly, SVM, neural networks and Decision Trees were exploited for Hungarian named entity classification at the phrase level ~\cite{farkasstatistical}. Semantic features and gazetteers have been used with the Bayesian network to recognize NEs of the Spanish language ~\cite{padro2005named}. Ando et al.~\cite{ando2005framework} used a structured machine learning algorithm for performing a multi-task learning based approach, where label prediction was considered as the primary task and masking of the current word was an auxiliary task. The classifier was selected based on the performance on the auxiliary task.
	
	A neural network architecture for NER was developed in 2008 by Collobert et al. ~\cite{collobert2008unified}, which relied on feature engineering, a dictionary, lexicon and orthographic features. Later, this architecture was modified with automatic feature extraction (at the level of word embedding) instead of using a feature engineered method ~\cite{collobert2011natural}.
	
	Deep learning models take input with units as words, characters, affixes and combinations of them, or even bytes. The Collobert et al. (2011)~\cite{collobert2011natural} model comprised of word-based features that are passed to the CRF layer via a convolutional layer. Later, the same model was enhanced by sequential features (an LSTM layer) on the English CONLL-2003 dataset~\cite{huang2015bidirectional}.
	
	Kim et al.~\cite{kim2016character} exploited character level features by generating word embeddings using bidirectional LSTM with Convolutional Neural Network (CNN) as a highway network. The predictions were made with softmax instead of CRF layer.
	
	Ma et al.~\cite{ma2016end} used a combination of characters and a word level representation and analyzed the impact on the Out-Of-Vocabulary (OOV) words, since the model does not perform well on OOV. Dernoncourt et al.~\cite{dernoncourt2017neuroner} followed the same model architecture to train the NER.
	Similarly, Santos et al.~\cite{santos2015boosting} obtained the word representation from characters by CNN and concatenated the embeddings of words before feeding to bidirectional LSTM. The Viterbi algorithm has also been used for inferencing for NER.
	
	Bharadwaj et al.~\cite{bharadwaj2016phonologically} has extended the model by the integration of phoneme as an additional feature. Similarly, Yadav et al.~\cite{yadav2018deep} integrated $n$-gram based most frequent affixes with the concatenated word representation for the same model for NER.
	
	\subsection{NER work so far on Indian Languages}
	
	One of the earliest works on NER for Indian languages was reported by Cucerzan et al.~\cite{cucerzan1999language}, mainly for Hindi. The author used a bootstrapping algorithm and an iterative learning algorithm to classify the names (both first and last name) and places on the 18806 tokens and achieved 41.70\% and 79.04\% F$_1$-score and accuracy, respectively. The IJCNLP 2008 workshop on NER for South and South East Asian Languages~\footnote{http://ltrc.iiit.ac.in/ner-ssea-08/} was the first shared task for NER for Indian (or South Asian) languages and it also reported perhaps the first dataset (in the public domain) for NER for Indian languages. It included five Indian languages: Hindi, Urdu, Bengali, Oriya and Telugu~\cite{singh2008named}. Of these, the Bengali dataset was developed by Jadavpur University and IIIT, Hyderabad, Urdu  by CRULP, Lahore and IIIT, Allahabad and the rest was developed by workshop organising institute (IIIT, Hyderabad). These datasets were annotated with 12 NE tags. Ekbal et al.~\cite{ekbal2008development} achieved 91.8\% as the best F$_1$-score from Support Vector Machines (SVM) on the annotated Bengali news corpus of 467858 tokens by 16 entities.
	
	
	Saha et al.~\cite{saha-etal-2008-hybrid} worked on Hindi NER by using Maximum Entropy Model (MaxEnt), which assigns an outcome for each token based on its history and features. They used about 243K words for training purposes, which was taken from the Dainik Jagran~\footnote{\url{https://www.jagran.com/}} (a popular Hindi newspaper), out of which about 16482 belonged to 4 named entities. Their MaxEnt based NER system was able to achieve a F$_1$-score of 81.52\%, using a hybrid set of Gazetteer, patterns, and lexical and contextual features.
	
	
	Sudha et al.~\cite{morwal2012named} worked on NER using Hidden Markov Model (HMM) for Hindi, Urdu and Punjabi. They used different number of tags for different corpora belonging to different domains. For example, they used Person, Location, River and Country tags in the tourism corpus; and Person, Time, Month, Dry-fruits and Food items tags in the story corpus.
	
	The Punjabi language is mainly written in two scripts: Gurmukhi (of Brahmi origin, like Devanagari) and Shahmukhi (a variant of the Persio-Arabic script). Most of the earlier works~\cite{kaur2009named,KAUR2015159,morwal2012named} on NER for this language were on data in Gurmukhi script and used statistical algorithms. Recent work on NER for Punjabi language using Shahamukhi script was explored by using 318275 tokens, out of which 16300 are entities such as Person, Location and Organization. The authors obtained 85.2\% as best F$_1$-score after applying a Recurrent Neural Network (RNN) over other classical and neural network techniques~\cite{Tayyab2020ner}.
	
	There is very little work on the Sambalpuri language, an Indo-Aryan language spoken in parts of the Indian state of Odisha. Behera et al. (2017) worked on Sambalpuri and Odia NER using SVM~\cite{behera1}. They took 112K words for Sambalpuri and 250K words for Odia. They made 7 labels for annotating named entities. The F$_1$-score measure obtained for Sambalpuri was 96.72\% and 98.10\% for Odia.

	Lalitha Devi et al. (in 2008) worked on 94K words of Tourism domain for Tamil NER by using CRF. They used a total of 106 tags divided into three categories of ENAMEX~\footnote{\url{http://cs.nyu.edu/cs/faculty/grishman/NEtask20.book_6.html\#HEADING17}}, TIMEX and NUMEX. They obtained 80.44\% F$_1$-score ~\cite{r-l-2008-domain}. Malarkodi C. S. et al. (2012)~\cite{c-s-etal-2012-tamil} also worked on Tamil NER using Conditional Random Field (CRF) model. They observed challenges in NER which occur due to several factors and are also applicable to other Indian languages as follows: agglutination, ambiguity, nested entities, spelling variations, name variations and the lack of capitalization.

	Rao et al.(2015)~\cite{rao2015esm} conducted a shared task as part of the FIRE 2015 conference on Entity Extraction From Social Media using Named Entity Recognizer for Indian languages. They collected their corpus using the Twitter API in the period of May-June 2015 for training data and August-September for testing data of Tamil, Malayalam, Hindi and English languages. They used 22 tags for different kinds of names. Different participating teams used various machine learning methods (CRF, SVM, HMM, Naive Bayes, Decision Tree) with diverse sets of features. The baseline F$_1$-score 47.10 for Hindi, 19.05 for Tamil 31.24 for Malayalam and 40.56 for English was reported.
	
	Ali et al.(2020)~\cite{ali2020siner} has reported very recent work on the NER corpus for Sindhi. In which they annotated over 1.35 million tokens with eleven entity classes using corpus derived from Awami Awaz and Kavish newspapers and reported a best benchmark F1-score of 89.16\% by using CNN-LSTM-CRF model on it.
	
	\section{Difficulties with Bhojpuri, Maithili, Magahi}
	
	Like many other modern Indo-Aryan languages, Bhojpuri, Maithili and Magahi are also non-tonal languages. They have Subject-Object-Verb (SOV) word order. Word-formation in these languages is somewhere between synthetic and analytical typology. There are a number of challenges while creating the corpus of these less-resourced languages. Despite being labelled as dialects of Hindi, morphological constructions of Bhojpuri, Maithili and Magahi considerably differ from that of Hindi. Linguistic differences and lexical ambiguity of these languages create challenges for machine learning and are responsible for many problems in Named Entity Recognition (NER).
	
	\subsection{Morphologically Rich Languages}
	
	Also like many other Indian languages, Bhojpuri, Maithili and Magahi are morphologically rich, so the identification of the `root' or lemma is challenging for these languages. They are partially synthetic languages. Hence, the use of embedded case markers, emphatic markers, classifiers, determiners etc. is frequent in these languages. These markers are responsible for many challenges in NER. Some examples are: \textit{mEWilIka}~\footnote{The example words are written in the WX notation} (Maithili's), \textit{Garasaz} (from home: -saz), \textit{jAwika} (caste of: -ka), \textit{gAmaka} (of village: -ka), \textit{gAmakez} (of village: -kez), \textit{rAmanAmIka} (name called Ram: -Ika), \textit{mircAI} (chillies: -AI) are from Maithili, \textit{SakunwaloM} (Sankuntla too/also: -oM), \textit{majaXAre} (in dillema: -e), \textit{baniyavoM} (seller too: -oM), \textit{GarahUz} (home as well: -Uz), \textit{Kewavo} (field too: -vo), \textit{surenxaravo} (surendra too: -vo), \textit{ekke} (one too: -ke; an example of gemination), \textit{sistAme} (in the system: -Ame), \textit{sahebAina} (feminine of Sahib, Memsahib: -Aina) in Bhojpuri and , \textit{rAwe} (in the night: -e), \textit{xuariyA} (door: -iyA), \textit{sonalo} (sonal too: -o), \textit{saberahIM} (in morning: -hIM), \textit{bABano} (Brahmin too: -o), \textit{Gare} (home too: -e) in Magahi. As can be seen from some of these examples, names can also be inflected in the three languages, creating problems for the algorithm. Some other markers in these examples do not apply to names, but by their very frequent occurrence and by their appearance on too many words, they pose challenges for NER.
	
	\subsection{Ambiguity}
	
	Bhojpuri, Maithili and Magahi, like many non-standardised (or less-standardised) languages, also have lexical ambiguity in unusual abundance~\footnote{Compared to other languages with very high numbers of speakers, as these latter languages tend to reduce lexical ambiguity in written language through means like standardization.}. This applies to many other Indian languages. These ambiguous words look similar but their tags are varied, which may pose a sense of perplexity or confusion for the NER task, especially for machine learning, both for annotation and for machine learning. Similar to many other languages Bhojpuri, Maithili and Magahi also have surprising amount of ambiguity even among proper names. According to our analysis, some of the examples are mentioned here in two categories: the ambiguity between the common and proper noun, and the ambiguity within the class of proper nouns. 
	
	\begin{enumerate}
		\item \textbf{Ambiguity between Common and Proper Nouns}
		\begin{enumerate}
			\item [--] Bhojpuri: \\
			\textit{cunarI} is ``A type of cloth [Common Noun]'' and ``Name of a Person [Proper Noun]'' as well. \\
			\textit{GAGarA} is ``A type of cloth [Common Noun]'' and ``Name of a River [Proper Noun]'' as well. \\
			\textit{kisalaya} is ``Young shoot/Bud [Common Noun]'' and ``Name of a Person [Proper Noun]'' as well.
			\item [--] Maithili: \\
			\textit{xaNdaka} is ``(of) Punishment [Common Noun]'' and is ``Name of a Forest (vana) [Proper Noun]'' also.\\
			\textit{GUGaru} is ``Anklet bells [Common Noun]'' as well as ``Name of a Person [Proper Noun]''.
		\end{enumerate}
		\item \textbf{Ambiguity of a Proper Name}
		\begin{enumerate}
			\item [--] People vs. Months:\\
			\textit{kAwika} and \textit{sAvana} are ``Name of a Person'' and ``Month of Indian Calendar'' as well.
			
			\item [--] People vs. Locations:\\
			\textit{surEyA}, \textit{bEjanAWa} and \textit{kexAranAWa} are ``Name of a person'' and ``Name of a place'' as well.
			
			\item [--] People vs. Seasons:\\
			\textit{basanwa}, \textit{hemaMwa} and \textit{SiSira} are ``Name of a person'' and ``Name of a Season'' as well.
			
			\item [--] People vs. Companies:\\
			\textit{sUryA} and \textit{bimala} are ``Name of a Company'' and ``Name of a person'' as well. 
			
			\item [--] Place vs. Companies:\\
			\textit{gvAliyara} is ``Name of a Company'' and ``Name of a place''.
			
			\item [--] Compound Words:\\
			\textit{rAmakiSana} and \textit{harimohana} are ``Name of a Person'', while \textit{rAma}, \textit{kiSana}, \textit{hari} and \textit{mohana} are ``Names of a Hindu God/person''.
		\end{enumerate}
	\end{enumerate}
	
	\subsection{Spelling Variations}
	
	Like many other Indian (less-standardised) languages, Bhojpuri, Maithili and Magahi languages also have the problem of spelling variation. In Bhojpuri, Maithili and Magahi speech communities, different people spell the same words differently. Because of this, a number of spelling variations of a single word create confusion and problems for the NER task. For example, \textit{laikiyA}, \textit{laikiA}, \textit{laikivA}, \textit{laikaniyA}, \textit{laikaniA} are variations of Boy and \textit{surasawiyA}, \textit{sArosawiyA}, \textit{sarasvawiyA} are variations of person name. These are not the usual inflections as given in examples in section 3.1, but represent features like familiarity or informality or deprecative usage.
	
	\subsection{Other Challenges}
	
	Whether it is a matter of traditional resources such as grammar books, dictionaries, textbooks, magazines, newspapers etc. or modern resources such as websites, blogs, emails, chats have increased in recent decades. Yet, compared to the requirements for successful NLP, especially data-driven NLP, the amount of linguistic resources (even simple text corpora) are still not available in sufficient quantities or with  for these languages due to their social status as dialects of Hindi. Even though Maithili has attained the status of language, it is still as resource scarce as Bhojpuri and Magahi. The lack of standardization and formal or official usage is one of the major problems for them, which poses difficulty in NER. 
	
	Apart from the scarcity of annotated resources, Bhojpuri, Maithili and Magahi also lack in terms of tools required for preprocessing such as Part-of -Speech tagging and Chunking that helps in recognizing NEs are not available. Or tools which are available have relatively poor performance so far.
	
	\section{Annotation}
	\label{annot}
	
	\subsection{NER Guidelines}
	
	We considered Indian Language Named Entities Tagset and Annotation Guidelines which were prepared under the Development of Cross Lingual Information Access (CLIA) system Phase-II Consortium Project funded by Ministry Communication and Information Technology (MCIT), Department of Information Technology, Government of India Version: 1.1~\footnote{\url{http://tdil-dc.in/index.php?option=com_download&task=showresourceDetails&toolid=815&lang=en}}. In this tagset, there are three main categories, viz. ENAMEX, NUMEX and TIMEX. There are a total of 22 tags in ENAMEX, NUMEX and TIMEX combined. In ENAMEX, there are eleven NER tags, viz. Person, Organization, Location, Facilities, Locomotives, Artifacts, Entertainment, Materials, Organisms, Plants and Disease. NUMEX consists of Distance, Money, Quantity and Count. TIMEX includes Time, Year, Month, Day, Date, Period and Special\_Day.
	
	\subsection{About Corpus}
	
	For this purpose, we considered the BMM corpus~\cite{mundotiya2020basic} of the project on Bhojpuri, Maithili, Magahi to Hindi Machine Translation System under Project Varanasi. The main goal of this project was to develop MT systems from Bhojpuri, Maithili, Magahi to Hindi. We have considered 16492, 9815 and 5320 sentences from the BMM corpus to create the NER annotated data. These sentences have 228373, 157468 and 56190 tokens and 32091, 23338 and 10175 types, for respective languages. After annotation, 12351 named entities in Bhojpuri, 19809 in Maithili and 7152 in Magahi were encountered, as mentioned in Table~\ref{lang_sent_ner_statistics}.
	
	\begin{table}[!h]
		\caption{Language-wise dataset statistics used for annotation of named entities}
		\centering
		\begin{tabular}{|c|ccccc|}
			\hline
			\bf Lang & \bf \#Sentences & \bf \#Tokens & \bf \#Types & \bf \#Entities & \bf \#Others\\
			\hline
			Bhojpuri & 16492 & 228373 & 32091 & 12351 & 216022 \\
			Maithili & 9815 & 157468 & 23338 & 19809 & 137659 \\
			Magahi & 5320 & 56190 & 10175 & 7152 & 49038 \\
			\hline
		\end{tabular}
		\label{lang_sent_ner_statistics}
	\end{table}
	
	\subsection{BMM NER Annotation}
	
	On the basis of the above mentioned guidelines the considered corpora were tagged with hierarchical tags. These tags were created considering Hindi as a sample language. For BMM, we mainly follow the guidelines used for Hindi for the purpose of NE annotation.  The broad categories and their statistics for each language are outlined in Table~\ref{lang_ner_statistics}. And each category with their statistics of hierarchical entities are summarised in Table~\ref{fine_ner_stat}. \\
	
	\begin{table}[!htbp]
		\caption{The statistics of annotated for the three broad categories for Bhojpuri, Maithili and Magahi}
		\centering
		\begin{tabular}{|c|ccc|}
			\hline
			\bf Lang & \bf ENAMEX & \bf NUMEX & \bf TIMEX \\
			\hline
			Bhojpuri & 10504 & 1152 & 695  \\
			Maithili & 15861 & 2214 & 1734 \\
			Magahi & 5790 & 725 & 637 \\
			\hline
		\end{tabular}
		\label{lang_ner_statistics}
	\end{table}
	
	\begin{table}[!h]
		\small
		\centering
		\caption{The statistics of annotated hierarchical entities for Bhojpuri, Maithili and Magahi. The ENAMEX, NUMEX and TIMEX categories contained 11, 4 and 7 hierarchical named entities. `Other' denotes regular words or tokens which are not named entities.}
		\label{fine_ner_stat}
		\begin{tabular}{|l|l|l|l|}
			\hline
			\bf Entities & \bf Bhojpuri & \bf Maithili & \bf Magahi \\
			\hline
			\multicolumn{1}{|c|}{\bf ENAMEX} &  &  &  \\
			Artifact  & 635 & 752 & 638 \\
			Disease  & 34 & 9 & 18 \\
			Entertainment  & 347 & 532 & 31 \\
			Facility  & 121 & 784 & 123 \\
			Location  & 985 & 4330 & 763 \\
			Locomotive  & 112 & 157 & 58 \\
			Material  & 278 & 481 & 379 \\
			Organism  & 481 & 222 & 566 \\
			Organization  & 109 & 2081 & 20 \\
			Person  & 7244 & 6462 & 3145 \\
			Plant  & 158 & 51 & 49 \\ \hline
			\multicolumn{1}{|c|}{\bf NUMEX} &  &  &  \\
			Count  & 685 & 1797 & 558 \\
			Distance  & 14 & 24 & 2 \\
			Money  & 166 & 131 & 112 \\
			Quantity  & 287 & 262 & 53 \\ \hline
			\multicolumn{1}{|c|}{\bf TIMEX} &  &  &  \\
			Date  & 69 & 48 & 5 \\
			Day  & 36 & 99 & 169 \\
			Month  & 66 & 281 & 52 \\
			Period  & 279 & 491 & 28 \\
			Special\_Day  & 8 & 210 & 1 \\
			Time  & 175 & 413 & 337 \\
			Year  & 62 & 192 & 45 \\ \hline
			\multicolumn{1}{|c|}{\bf OTHER} &  &  &  \\
			Other & 216022 & 137659 & 49038 \\
			\hline
		\end{tabular}
	\end{table}
	
	\section{Algorithms}
	\label{algorithm}
	
	For performing the baseline experiments on the prepared NER dataset of Bhojpuri, Maithili and Magahi, we have used two standard techniques which are known to provide previous state-of-the-art results for NER for other languages. This techniques are: Conditional Random Fields (CRF)~\cite{lafferty2001conditional}, a statistical algorithm and LSTM-CNNs-CRF~\cite{ma-hovy-2016-end}, based on a deep learning algorithm. From our study on related work of NER, a statistical algorithm (CRF) yielded considerable comparative result to the Deep Learning method, perhaps because of the lack of sufficient data.
	
	\subsection{Conventional Machine Learning Algorithm}
	
	Conditional Random Field (CRF) is a discriminative model that uses conditional probability and is well suited for performing sequential prediction. Coined as CRF by Lafferty et al. \cite{lafferty2001conditional,rozenfeld2006systematic}, this undirected graphical model learns the dependency between each state and the entire input sequence. An input of each feature function has multiple input values as the sentence, word position of the sentence, and label of the current, and previous word that provides the contextual information. The feature function is expected to express some kind of characteristic of the sequence.  A set of weights are assigned (initializing to random values) to the feature function to build the conditional field. Maximum Likelihood Estimation estimates the parameters. Finally, gradient descent updates the parameter values iteratively until the values converge.
	
	\subsection{Deep Learning Algorithm}
	
\begin{figure}[!h]
		\centering
		\includegraphics[width=0.7\linewidth]{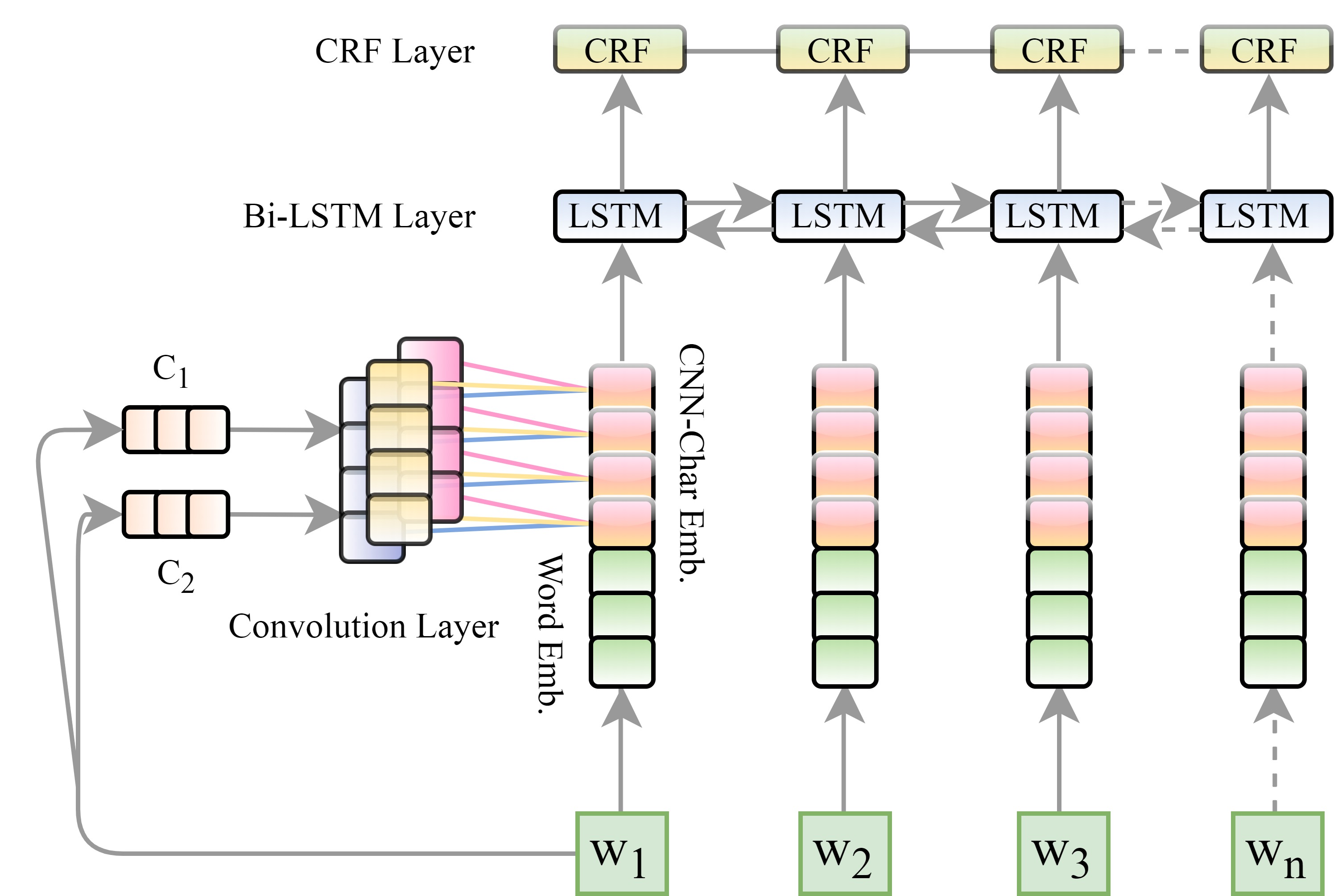}
		\caption{LSTM-CNNs-CRF architecture, where $c_1$, $c_2$ are character representations and $w_n$ is word representation. The $Char\_emb$ and $e_{1:n}$ are denoted by CNN-Char Emb. and Word Emb. respectively.}
		\label{fig:cnn-rnn-crf}
	\end{figure}

	LSTM-CNNs-CRF model~\cite{ma-hovy-2016-end} consists of three components which are sequentially arranged. These components: CNN, LSTM and CRF. They are responsible for capturing the character-level information, word-level information and the dependency information. The architecture is depicted in Figure~\ref{fig:cnn-rnn-crf}.
	\\
	\\
	\textbf{The CNN Layer}\\
	The CNN model helps to extract character $n$-gram information of the given input words for our problem. It has three essential operations, which are Convolution, Pooling and Feed-forward. The convolution layer performs convolution operations in which filters are applied over the fixed-length sequential input, where each filter has a certain window size to produce a new feature.
	
	Let an input word $W_1$ has $c_1, \ldots, c_m$ characters, encoded by a one-hot vector of $o_1, \ldots, o_m$, over which $k$ filters have been applied with $h$ window size so that the generated feature becomes:
	\begin{equation}
		C = [C_1, C_2, \ldots, C_{m-h+1}]
	\end{equation}
	\begin{equation}
		where, C_i = ReLU[Fo_{i:i+h-1}+b]
	\end{equation}
	
	Here, $C$ is feature map produced by $k$ filters. The Pooling operation is performed by the max-pooling layer on the feature maps to extract the most relevant information to get a word vector. 
	\begin{equation}
		\hat{C} = \max\{C\}
	\end{equation}
	
	These obtained features $\{\hat{C_1}, \hat{C_2}, \hat{C_3}, \ldots, \hat{C_k} \}$ are fed to the fully connected layer for obtaining the desired vector size of a word from characters.
	\begin{equation}
		Char\_emb_{1} = W^T.\hat{C}_{1:k} + b
	\end{equation}
	
	The input layer at the word-level encodes the word into a fixed-length real-valued vector ($e$), which is learnt during the model training. The final word representation is obtained after combining the actual word representation and character-level word representation. These encoded-word vector and character-level word vector are concatenated as the final word embedding ($WE$).
	\begin{equation}
		WE_{1:n} = Char\_emb_{1:n} \oplus e_{1:n}
	\end{equation}
	\\
	\textbf{The LSTM Layer}
	\\
	The final word embeddings are then passed to the LSTM layer for capturing the longer dependencies among the words of the input sentence. BMM languages tend to have long-distance dependencies. Here, the bidirectional LSTM layer has been used for modelling the longer dependencies by applying LSTM on the forward and the backward directions. The obtained hidden states from both directions is concatenated and is fed to the next layer:
	\begin{equation}
		\overrightarrow{h_{1:n}} = \overrightarrow{LSTM}(WE_{1:n})
	\end{equation}
	\begin{equation}
		\overleftarrow{h_{1:n}} = \overleftarrow{LSTM}(WE_{1:n})
	\end{equation}
	\begin{equation}
		h = \overrightarrow{h_{1:n}} \oplus \overleftarrow{h_{1:n}}
	\end{equation}
	\\
	\textbf{The CRF Layer} \\
	After capturing dependencies among the words by the LSTM layer, a CRF layer is used to capture dependencies of the labels. The CRF layer is applied over the representations obtained from the bidirectional LSTM output for obtaining the label dependencies of the input sentences. Linear CRF models the linear relationship with previous labels to generate the probability score of the current labels by the potential function $\psi(.)$:
	\begin{equation}
		\psi(h_i, y_i, y_{i-1}) = \exp{(y_i^T W_1^T h_i + y_{i-1}^T W_2 y_i)}
	\end{equation}
	\begin{equation}
		p(y|h;\theta) = \frac{\Pi_{i=1}^{n} \psi(h_i, y_i, y_{i-1})}{\Sigma_{y'\in Y} \Pi_{i=1}^{n} \psi(h_i, y'_i, y'_{i-1})}
	\end{equation}
	
	Here, $Y$ denotes all possible labels and $\theta$ is the learning parameter.
	
	\section{Experiments}
	\label{experiment}
	
	The dataset was divided into training and testing splits with a ratio of 80-20. While splitting, it was ensured that the testing dataset included all the named entities with frequency of at least one. The dataset statistics after splitting are shown in the Table~\ref{train-test-stat}. The training strategy and the obtained results have been explained in the following sections.
	
	\begin{table}[!h]
		\caption{The dataset sizes for each language. The out-of-vocabulary (OOV) percentage is calculated by token-type differences between test data and the training data.}
		\label{train-test-stat}
		\begin{tabular}{|c|c|l|l|l|l|}
			\hline
			\multicolumn{1}{|l|}{\textbf{Language}} & \multicolumn{1}{l|}{\textbf{Data-Mode}} & \textbf{Sentences} & \textbf{Tokens} & \textbf{Types} & \textbf{OOV (\%)} \\ \hline
			\hline
			\multirow{2}{*}{\textbf{Bhojpuri}} & Train & 11544 & 160226 & 19642 & \multirow{2}{*}{25.50} \\ \cline{2-5}
			& Test  & 4948  & 68147  & 12449 &                        \\ \hline
			\multirow{2}{*}{\textbf{Maithili}} & Train & 7849  & 125442 & 15859 & \multirow{2}{*}{27.94} \\ \cline{2-5}
			& Test  & 1966  & 32026  & 7479  &                        \\ \hline
			\multirow{2}{*}{\textbf{Magahi}}   & Train & 4256  & 44833  & 6868  & \multirow{2}{*}{22.34} \\ \cline{2-5}
			& Test  & 1065  & 11357  & 3307  &                        \\
			\hline	
		\end{tabular}
	\end{table}
	
	\subsection{Training the CRF Model}
	
	There are several implementations of CRF that are publicly available. We have used the CRFsuit~\footnote{https://sklearn-crfsuite.readthedocs.io/en/latest/} implementation with the training algorithm of L-BFGS, executed up to a maximum of 100 iterations. To avoid overfitting and underfitting issues of CRF, C1 and C2 regularization parameters with random search cross-validation have been used for training, where the value of cross-validation is 3, and the number iterations are 50. The current word, the neighbouring words with the adjacency of 2, affixes of the current word with a window size of 3, whether the current word and the neighbouring words are digits and whether the current word is first, or the last word of a sentence are considered as hand-crafted features for training the CRF model. The optimal values of C1 and C2 are 0.178 and 0.006 for Magahi, 0.440 and 0.018 for Maithili and 0.481 and 0.003 for Bhojpuri as obtained after training.
	
	\subsection{Training the LSTM-CNNs-CRF Model}
	
	
	The LSTM-CNNs-CRF model takes input in the form of characters and words to generate word embedding to overcome the scarcity of annotated data. As Deep Learning models are very sensitive to the the data size as well as the values of the parameters, it is not guaranteed that the same value of the parameter will provide optimal results for another language. The word embeddings, character embeddings and the hidden representations plays a vital role in obtaining the best possible results. For our experiments, the sizes of the word embedding size, the character embedding and the hidden representations 100, 20 and 25, respectively for Bhojpuri. Similarly, the values are 100, 30 and 50 for Magahi and 200, 20 and 50 for Maithili. The number of convolutional layers is 4. The model training is performed with the Stochastic Gradient Descent (SGD) optimizer, where the learning rate is 0.015, which decays over the epoch by 0.05 for constraint learning. During training, the L2 regularizer and dropout are also used with the values of 0.5 and $1\mathrm{e}{-8}$, respectively to prevent model overfitting. The summary of the (hyper-)parameters with their values are mentioned in Table ~\ref{tab:para_hyperpara}.
	
	\begin{table}[!h]
		\caption{The value of (hyper-)parameters used for training of the LSTM-CNNs-CRF model}
		\label{tab:para_hyperpara}
		\centering
		\begin{tabular}{|l|l||l|l|}
			\hline
			\textbf{(Hyper-)Parameter} & \textbf{Value} & \textbf{(Hyper-)Parameter} & \textbf{Value} \\
			\hline
			Word Embedding & [100, 200] & Word Hidden & 200 \\
			Char. Embedding & [20, 30] & Char. Hidden & [25, 50] \\
			Batch Size & 20 & Epochs & 20 \\
			Convolution Layer & 4 & Optimizer & SGD \\
			Dropout & 0.5 & L2 & 1e-8 \\
			Learning Rate & 0.015 & Learning Decay & 0.05 \\
			\hline
		\end{tabular}
	\end{table}

	\section{Results}
	
	Bhojpuri has a larger annotated dataset than the remaining two languages. For this language, the obtained results on the validation data for CRF and LSTM-CNNs-CRF are 70.56\% and 61.41\% as F$_1$-score, respectively. The entity-wise Precision, Recall and F$_1$-score given in Table~\ref{tab:bhoj_score} shows that LSTM-CNNs-CRF struggles to learn low-frequency entities such as Day, Disease, Distance, Organization, Special\_Day and Year. 
	
	In the CRF training, the best transition is obtained from I-Money$\to$ I-Money, B-Organization$\to$I-Organization, B-Period$\to$I-Period, B-Facility$\to$I-Facility and B-Year$\to$I-Year. Similarly the worst transition represents the wrong interpretation as B-Count$\to$I-Person, B-Count$\to$B-Count, I-Person$\to$B-Artifact and B-Artifact$\to$B-Artifact.

	\begin{table}[!h]
		\small
		\caption{NER tag-wise scores obtained by CRF and LSTM-CNNs-CRF for Bhojpuri. The metrics, which are \textbf{P}recision, \textbf{R}ecall and \textbf{F$_1$}-score}
		\centering
		\begin{tabular}{|l|l|l|l||l|l|l|}
			\hline
			\textbf{Techniques} & \multicolumn{3}{c||}{\textbf{LSTM-CNNs-CRF}} & \multicolumn{3}{c|}{\textbf{CRF}} \\ \hline
			
			\textbf{NER-Tag} & \textbf{P} & \textbf{R} & \textbf{F1} & \textbf{P} & \textbf{R}  & \textbf{F1} \\ \hline
			
			\hline
			Artifact & 84.91 & 26.16 & 40.00 & 96.77 & 34.88 & 51.28 \\
			Disease & 0.00 & 0.00 & 0.00 & 100.00 & 77.78 & 87.50 \\
			Entertainment & 77.78 & 6.93 & 12.73 & 84.00 & 20.79 & 33.33 \\
			Facility & 25.00 & 10.26 & 14.55 & 36.00 & 23.08 & 28.12 \\
			Location & 97.20 & 39.10 & 55.76 & 95.36 & 54.14 & 69.06 \\
			Locomotive & 65.00 & 32.50 & 43.33 & 73.91 & 42.50 & 53.97 \\
			Material & 77.78 & 7.37 & 13.46 & 91.67 & 23.16 & 36.97 \\
			Organism & 100.00 & 7.69 & 14.29 & 100.00 & 18.46 & 31.17 \\
			Organization & 0.00 & 0.00 & 0.00 & 100.00 & 18.18 & 30.77 \\
			Person & 98.88 & 72.45 & 83.62 & 98.97 & 79.01 & 87.87 \\
			Plant & 90.91 & 25.64 & 40.00 & 84.62 & 56.41 & 67.69 \\
			Count & 71.43 & 23.26 & 35.09 & 71.62 & 30.81 & 43.09 \\
			Distance & 0.00 & 0.00 & 0.00 & 50.00 & 20.00 & 28.57 \\
			Money & 100.00 & 6.67 & 12.50 & 88.46 & 38.33 & 53.49 \\
			Quantity & 55.56 & 14.08 & 22.47 & 48.78 & 28.17 & 35.71 \\
			Date & 100.00 & 5.88 & 11.11 & 100.00 & 23.53 & 38.10 \\
			Day & 0.00 & 0.00 & 0.00 & 100.00 & 16.67 & 28.57 \\
			Month & 0.00 & 0.00 & 0.00 & 100.00 & 27.78 & 43.48 \\
			Period & 100.00 & 14.47 & 25.29 & 86.67 & 17.11 & 28.57 \\
			Special\_Day & 0.00 & 0.00 & 0.00 & 0.00 & 0.00 & 0.00 \\
			Time & 100.00 & 28.57 & 44.44 & 100.00 & 26.79 & 42.25 \\
			Year & 0.00 & 0.00 & 0.00 & 57.14 & 40.00 & 47.06 \\ \hline
			\textbf{Avg. score} & \textbf{91.14} & \textbf{50.51} & \textbf{61.41} & \textbf{93.64} & \textbf{59.90} & \textbf{70.56} \\ \hline
			OTHER & 97.39 & 99.46 & 98.41 & 97.94 & 99.08 & 98.51 \\ \hline
			Avg. score & 96.15 & 96.84 & 96.25 & 96.59 & 96.99 & 96.73 \\
			(including OTHER) &  &  &  &  &  &  \\
			\hline
		\end{tabular}
		\label{tab:bhoj_score}
	\end{table}

	For Maithili, we obtained 73.19\% F$_1$-score for CRF and 71.38\% for LSTM. The tag-wise scores are listed in Table~\ref{tab:mai_score}.
	
	
	Some of the remaining entities (Day, Date, Month, Year, Distance, Organism and Plant) have rare intermediates. The optimal transitions from this language's annotated dataset are B-Entertainment$\to$I-Entertainment, B-Facility$\to$I-Facility and B-Organism$\to$I-Organism, and worst transitions are B-Period$\to$B-Count, B-Location$\to$I-Person and B-Location$\to$B-Period. 
	
	\begin{table}[!h]
		\caption{NER tag-wise scores obtained for CRF and LSTM-CNNs-CRF for Maithili}
		\centering
		\begin{tabular}{|l|l|l|l||l|l|l|}
			\hline
			\textbf{Techniques} & \multicolumn{3}{c||}{\textbf{LSTM-CNNs-CRF}} & \multicolumn{3}{c|}{\textbf{CRF}} \\ \hline
			
			\textbf{NER-Tag} & \textbf{P} & \textbf{R} & \textbf{F1} & \textbf{P} & \textbf{R}  & \textbf{F1} \\ \hline
			
			\hline
			Artifact & 83.33 & 28.09 & 42.02 & 77.78 & 31.46 & 44.80 \\
			Disease & 0.00 & 0.00 & 0.00 & 0.00 & 0.00 & 0.00 \\
			Entertainment & 70.18 & 68.97 & 69.57 & 87.80 & 62.07 & 72.73 \\
			Facility & 81.25 & 43.82 & 56.93 & 80.77 & 47.19 & 59.57 \\
			Location & 93.88 & 66.72 & 78.01 & 91.96 & 68.90 & 78.78 \\
			Locomotive & 100.00 & 15.15 & 26.32 & 93.75 & 45.45 & 61.22 \\
			Material & 88.57 & 39.74 & 54.87 & 85.71 & 38.46 & 53.10 \\
			Organism & 100.00 & 5.41 & 10.26 & 66.67 & 5.41 & 10.00 \\
			Organization & 92.31 & 57.93 & 71.19 & 93.64 & 55.86 & 69.98 \\
			Person & 97.08 & 70.89 & 81.94 & 97.11 & 75.62 & 85.03 \\
			Plant & 100.00 & 45.45 & 62.50 & 100.00 & 36.36 & 53.33 \\
			Count & 83.57 & 72.65 & 77.73 & 87.69 & 69.80 & 77.73 \\
			Distance & 0.00 & 0.00 & 0.00 & 50.00 & 33.33 & 40.00 \\
			Money & 66.67 & 15.38 & 25.00 & 62.50 & 38.46 & 47.62 \\
			Quantity & 77.78 & 11.11 & 19.44 & 83.33 & 7.94 & 14.49 \\
			Date & 100.00 & 20.00 & 33.33 & 100.00 & 60.00 & 75.00 \\
			Day & 69.23 & 45.00 & 54.55 & 85.71 & 30.00 & 44.44 \\
			Month & 100.00 & 88.37 & 93.83 & 97.44 & 88.37 & 92.68 \\
			Period & 92.16 & 63.51 & 75.20 & 90.32 & 75.68 & 82.35 \\
			Special\_Day & 100.00 & 62.07 & 76.60 & 95.00 & 65.52 & 77.55 \\
			Time & 83.33 & 24.59 & 37.97 & 88.89 & 26.23 & 40.51 \\
			Year & 100.00 & 66.67 & 80.00 & 100.00 & 59.26 & 74.42 \\ \hline
			\textbf{Avg. score} & \textbf{91.60}	& \textbf{60.65} &	\textbf{71.38}	& \textbf{91.53} &	\textbf{62.86} & \textbf{73.19} \\ \hline
			OTHER & 95.27 & 98.61 & 96.91 & 95.52 & 98.33 & 96.90 \\ \hline
			Avg. score & 93.34 & 93.85 & 93.33 & 93.32 & 93.87 & 93.33 \\
			(including OTHER) &  &  &  &  &  &  \\
			
			\hline
		\end{tabular}
		\label{tab:mai_score}
	\end{table}
	
	For Magahi, we obtained F$_1$-scores of 84.18\% and 86.39\% for CRF and LSTM-CNNs-CRF, respectively. The tag-wise scores are listed in Table~\ref{tab:mag_score}.
	
	
	
	\begin{table}[!h]
		\caption{NER tag-wise scores obtained for CRF and LSTM-CNNs-CRF for Magahi}
		\centering
		\begin{tabular}{|l|l|l|l||l|l|l|}
			\hline
			\textbf{Techniques} & \multicolumn{3}{c||}{\textbf{LSTM-CNNs-CRF}} & \multicolumn{3}{c|}{\textbf{CRF}} \\ \hline
			
			\textbf{NER-Tag} & \textbf{P} & \textbf{R} & \textbf{F1} & \textbf{P} & \textbf{R}  & \textbf{F1} \\ \hline
			
			\hline
			Artifact & 100.00 & 58.89 & 74.13 & 100.00 & 54.44 & 70.50 \\
			Disease & 0.00 & 0.00 & 0.00 & 0.00 & 0.00 & 0.00 \\
			Entertainment & 100.00 & 40.00 & 57.14 & 100.00 & 40.00 & 57.14 \\
			Facility & 100.00 & 72.22 & 83.87 & 100.00 & 72.22 & 83.87 \\
			Location & 96.83 & 70.11 & 81.33 & 98.33 & 67.82 & 80.27 \\
			Locomotive & 80.00 & 40.00 & 53.33 & 75.00 & 30.00 & 42.86 \\
			Material & 94.29 & 91.67 & 92.96 & 96.77 & 83.33 & 89.55 \\
			Organism & 97.96 & 77.42 & 86.49 & 100.00 & 75.81 & 86.24 \\
			Organization & 100.00 & 25.00 & 40.00 & 100.00 & 50.00 & 66.67 \\
			Person & 100.00 & 88.29 & 93.78 & 97.59 & 84.98 & 90.85 \\
			Plant & 100.00 & 62.50 & 76.92 & 100.00 & 62.50 & 76.92 \\
			Count & 95.65 & 85.71 & 90.41 & 96.67 & 75.32 & 84.67 \\
			Distance & 0.00 & 0.00 & 0.00 & 0.00 & 0.00 & 0.00 \\
			Money & 100.00 & 95.00 & 97.44 & 100.00 & 95.00 & 97.44 \\
			Quantity & 100.00 & 54.55 & 70.59 & 100.00 & 54.55 & 70.59 \\
			Date & 0.00 & 0.00 & 0.00 & 100.00 & 66.67 & 80.00 \\
			Day & 94.44 & 89.47 & 91.89 & 94.44 & 89.47 & 91.89 \\
			Month & 100.00 & 72.73 & 84.21 & 100.00 & 72.73 & 84.21 \\
			Period & 100.00 & 33.33 & 50.00 & 100.00 & 33.33 & 50.00 \\
			Special\_Day & 0.00 & 0.00 & 0.00 & 0.00 & 0.00 & 0.00 \\
			Time & 100.00 & 72.97 & 84.37 & 100.00 & 70.27 & 82.54 \\
			Year & 100.00 & 75.00 & 85.71 & 100.00 & 62.50 & 76.92 \\ \hline
			\textbf{Avg. score} & \textbf{97.82} & \textbf{78.42} & \textbf{86.39} & \textbf{97.67} & \textbf{75.00} & \textbf{84.18} \\ \hline
			OTHER & 96.92  & 98.28  & 97.60 & 96.44 & 98.43 & 97.42 \\ \hline
			Avg. score & 95.44 & 95.62 & 95.44 & 95.11 & 95.29 & 95.04 \\
			(including OTHER) &  &  &  &  &  &  \\
			\hline
		\end{tabular}
		\label{tab:mag_score}
	\end{table}
	
	
	
	
	\section{Conclusion}
	Bhojpuri, Maithili and Magahi are Purvanchal languages which are often considered  dialects of Hindi, even though they are widely spoken in parts of India. Bhojpuri is spoken even outside India. Partly due to their dialectal nature, they show more linguistic variations such as nominal case inflection, emphatic expressions. Like other computational resources, there is a lack of any NER system for these languages. We describe a first attempt at this. This attempt includes the creation of a dataset as well as reporting the results for two baseline systems, one that uses CRF and the other that uses an LSTM-CNNs-CRF model. These NER systems are planned to be used in machine translation system for Bhojpuri, Maithili and Magahi to Hindi. The NER dataset, prepared by native speaker linguists, consists of 228373, 157468 and 56190 tokens, out of which 12351, 19809 and 7152 are NE’s. The tagset used is a union of ENAMEX, TIMEX and NUMEX tagsets, having a total of 22 labels. The results obtained (in terms of F$_1$-score) are 70.56\% for 61.41\% for Bhojpuri with CRF and LSTM-CNNs-CRF, respectively. The results for Maithili are 73.19\% and 71.38\% and for Magahi, they are 84.18\% and 86.39\% for the two models. Even though the total data size is more for Bhojpuri, the scores are lower as the number of NEs in the dataset of this languages is relatively much less than for the other languages. In other words, the results are consistent with the number of NEs in the datasets, rather than with the total size of the dataset in number of tokens.

	\section*{Acknowledgements}
	We would like to thank our NER annotators Ajeet Kumar and Umesh Kumar Chaudhary for Bhojpuri, Nihal Kumar for Maithili, Rishav Raj Singh, Supriya Chauhan and Manu Mashani for Magahi. All the annotators are native speakers of the respective languages and did their Masters in Linguistics from Banaras Hindu University, Varanasi.

	\bibliography{BMM_NER_base}

	\appendix
	\section{}
	The following figures give the confusion matrices for the reported prediction experiments. While plotting a confusion matrix, we have ignored NEs that: (i)  have both actual and predicted counts as zeros, (ii) or are all predicted correctly. For example, Year, Special\_Day, Distance and Disease in Bhojpuri, and Distance and Special\_Day in Magahi belong to the case (i), whereas Time and Period in Bhojpuri, and Month, Time, Year, Person, Facility, Organization and Entertainment in Magahi, and Month in Maithili are all predicted correctly by the LSTM-CNNs-CRF model. 
	
	\begin{figure}
		\centering
		\subfigure{\label{fig:a}\includegraphics[width=0.9\linewidth]{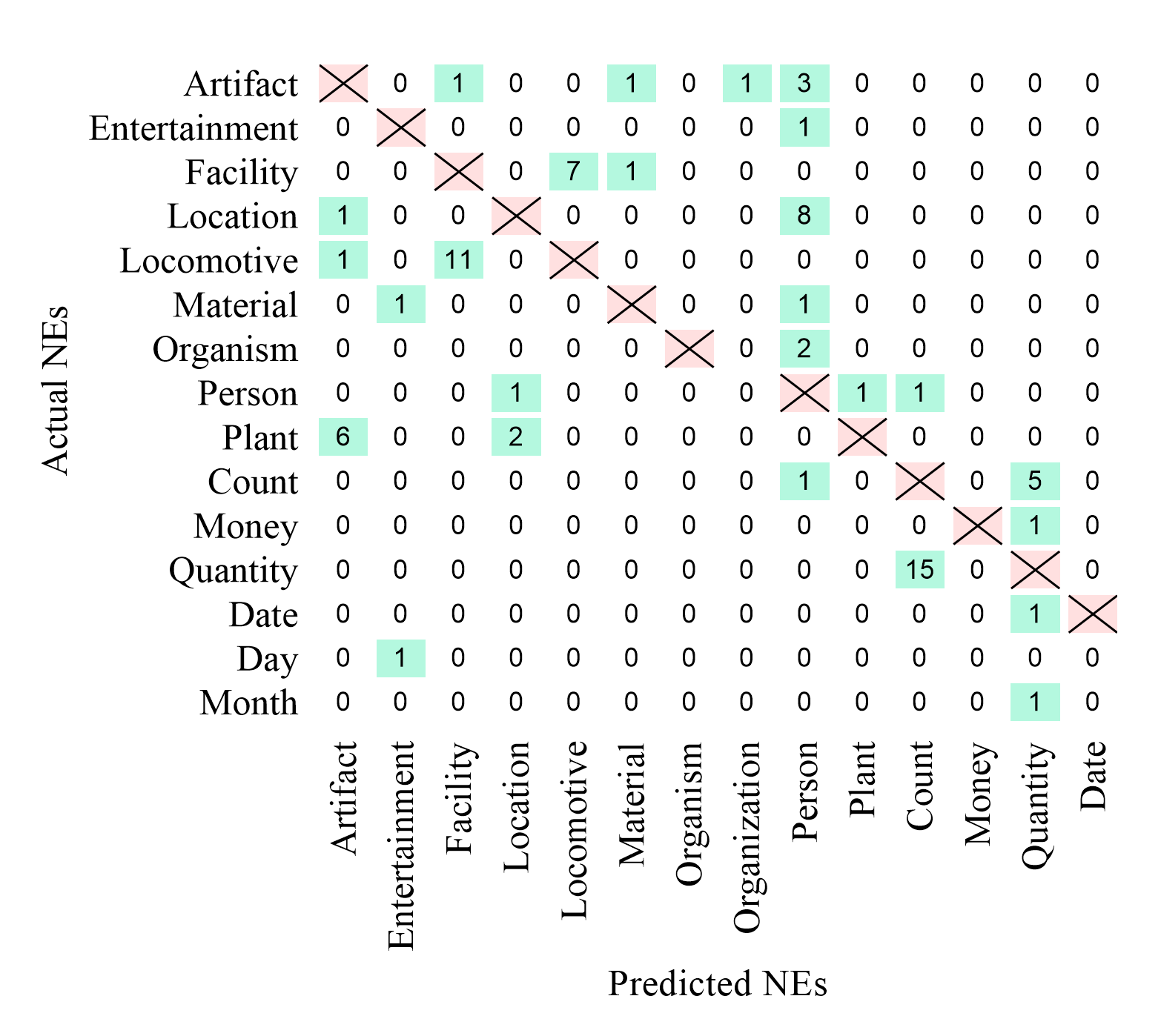}}
		\subfigure{\label{fig:b}\includegraphics[width=0.8\linewidth]{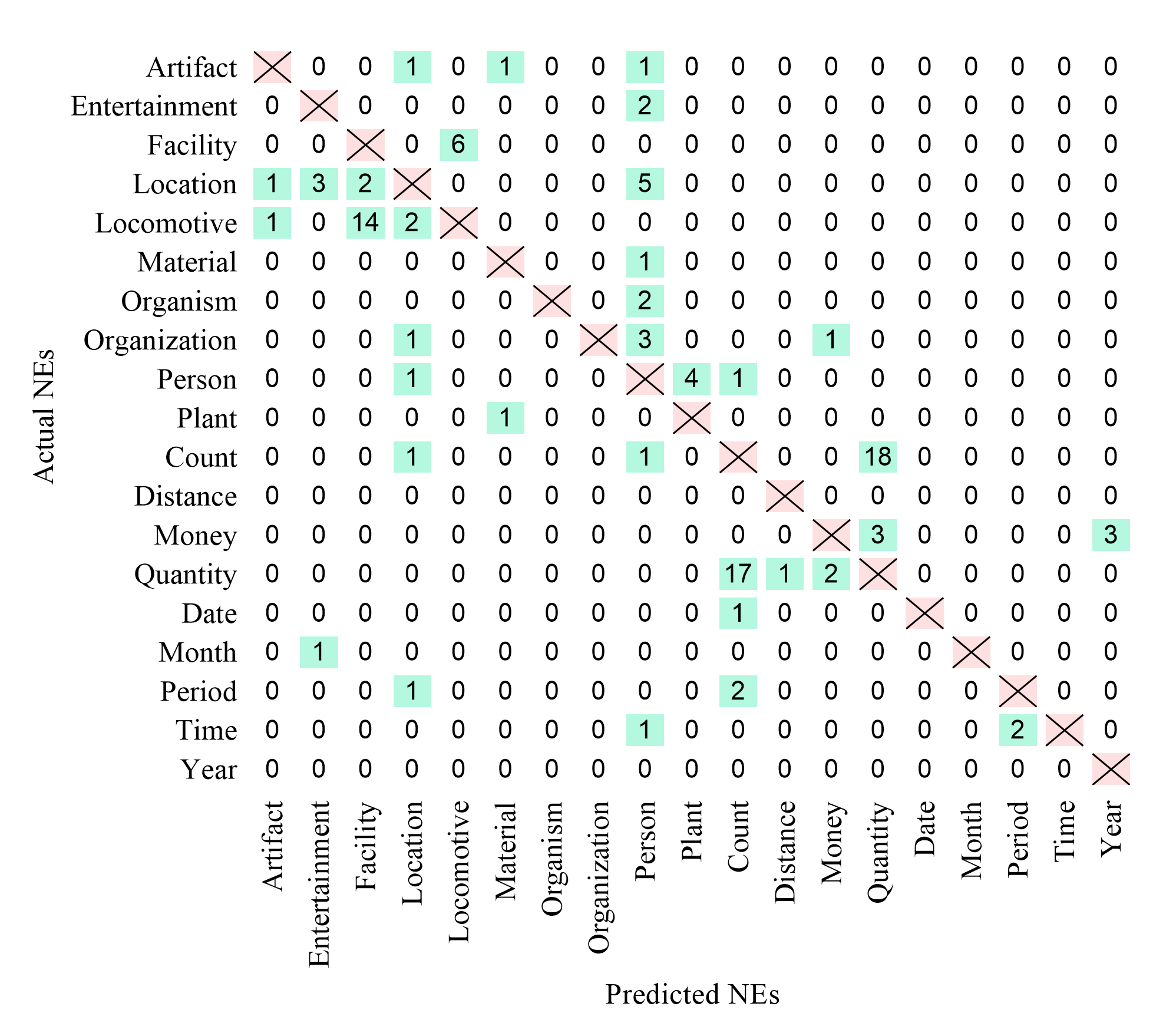}}
		\caption{Confusion matrix for Bhojpuri for the LSTM-CNNs-CRF (above) and CRF (below) models; The \colorbox{pink!70}{$\times$} refers to correctly prediction}
		\label{bhoj_cm_nn_crf}
	\end{figure}

	\begin{figure}
		\centering
		\subfigure{\includegraphics[width=0.7\textwidth]{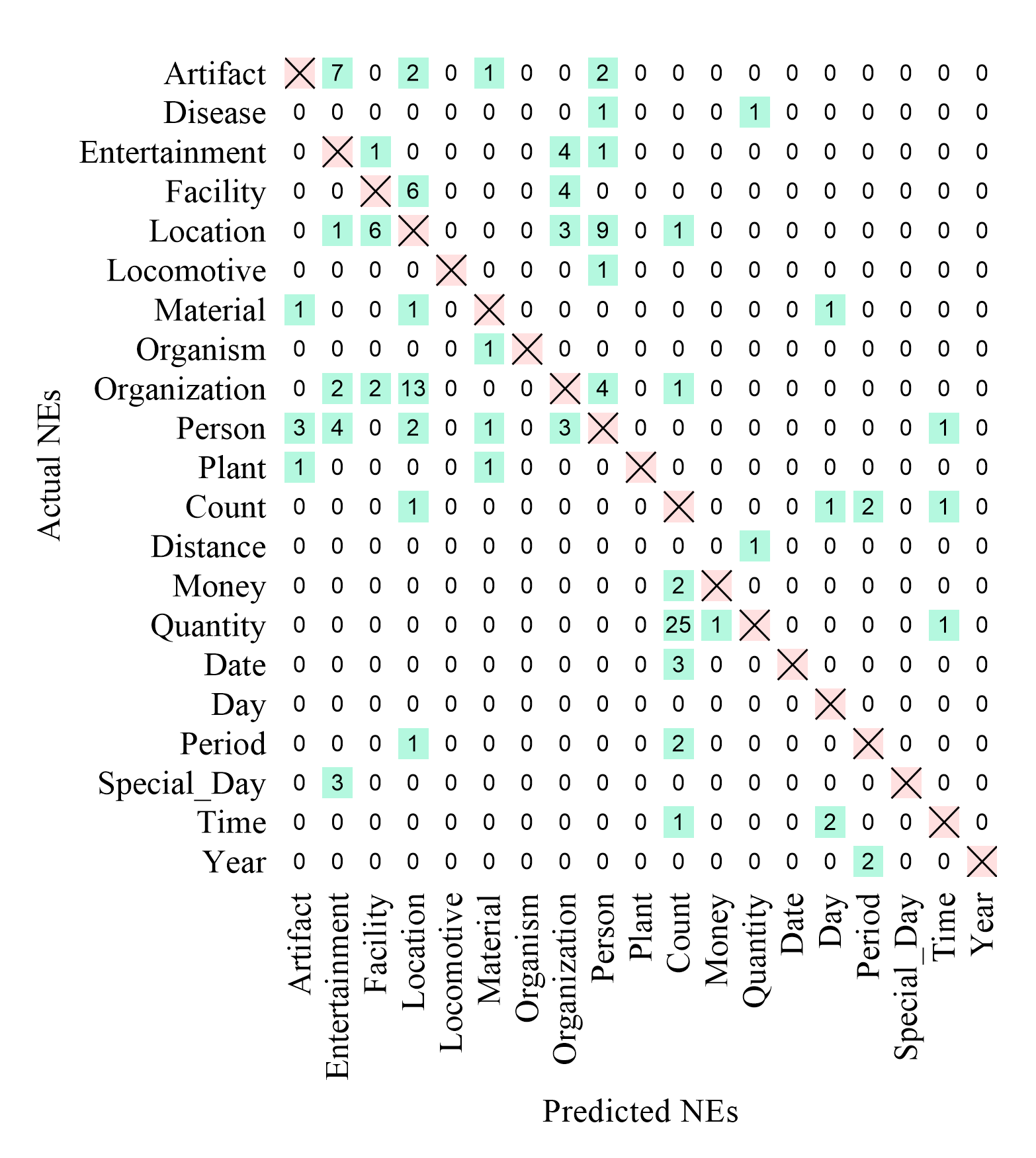}}
		\subfigure{\includegraphics[width=0.7\textwidth]{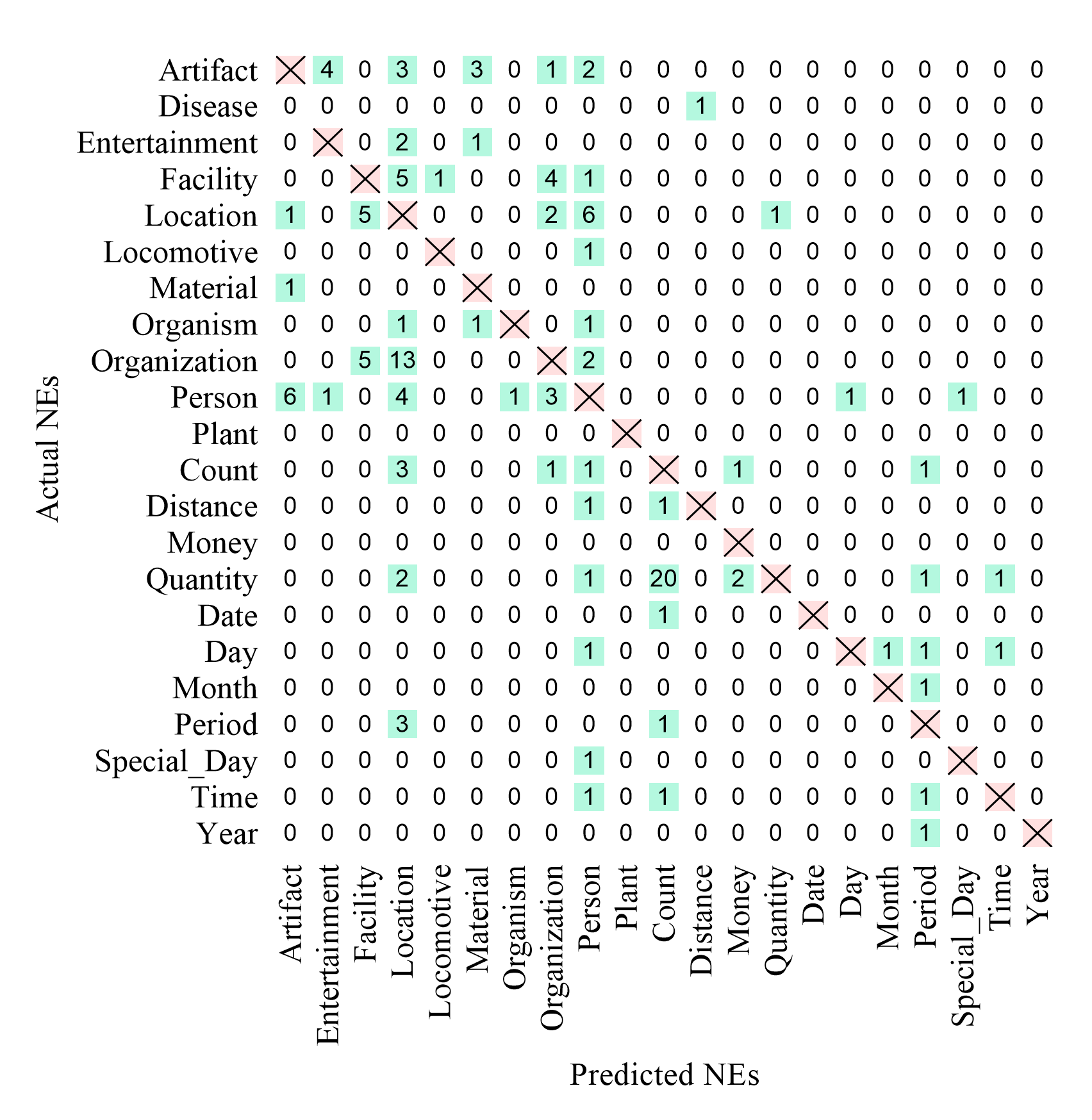}}
		\caption{Confusion matrix for Maithili for the LSTM-CNNs-CRF (above) and CRF (below) models; The \colorbox{pink!70}{$\times$} refers to correctly prediction}
		\label{mai_cm_nn_crf}
	\end{figure}

	\begin{figure}
		\centering
		\subfigure{\includegraphics[width=0.8\textwidth]{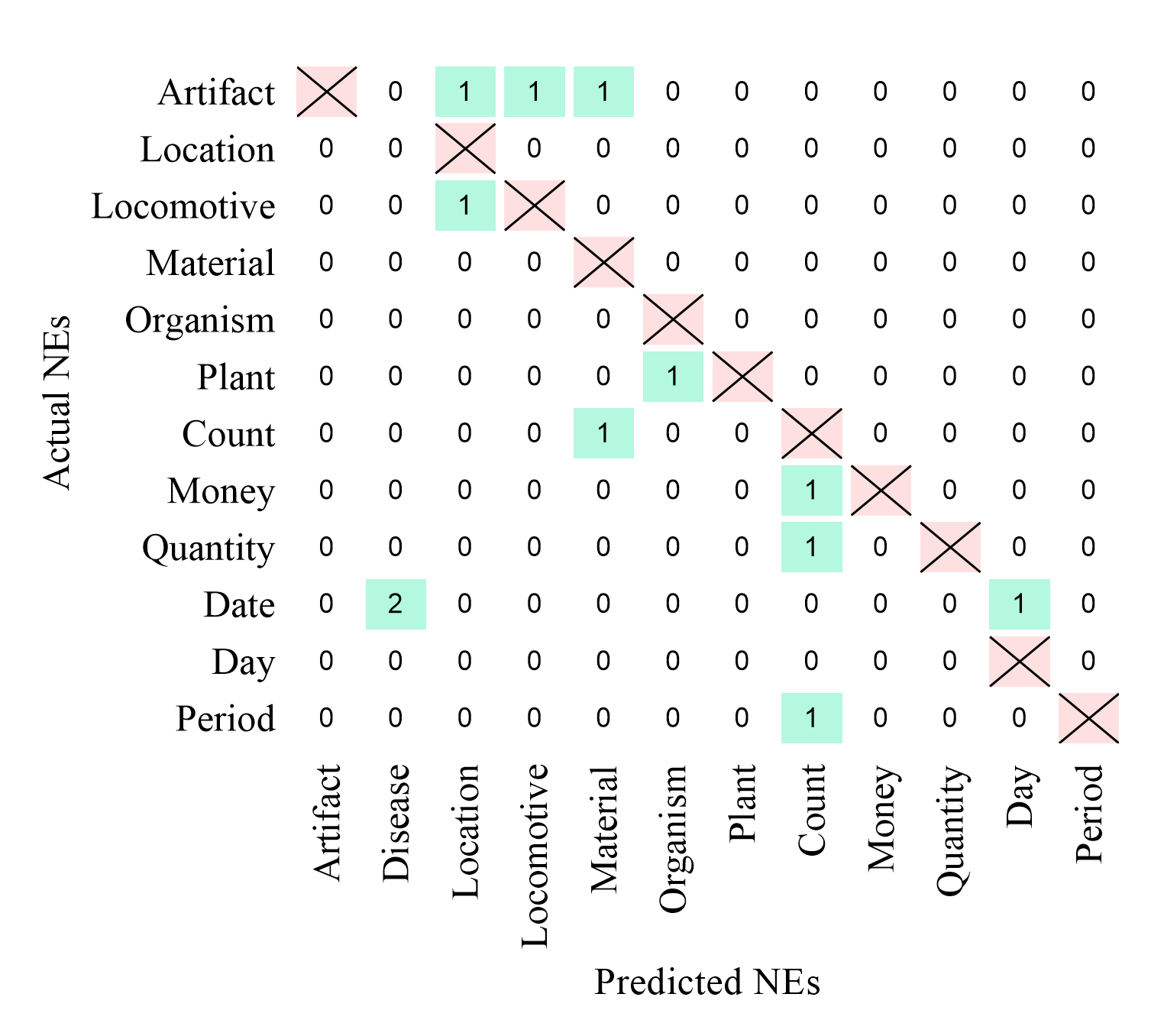}}
		\subfigure{\includegraphics[width=0.8\textwidth]{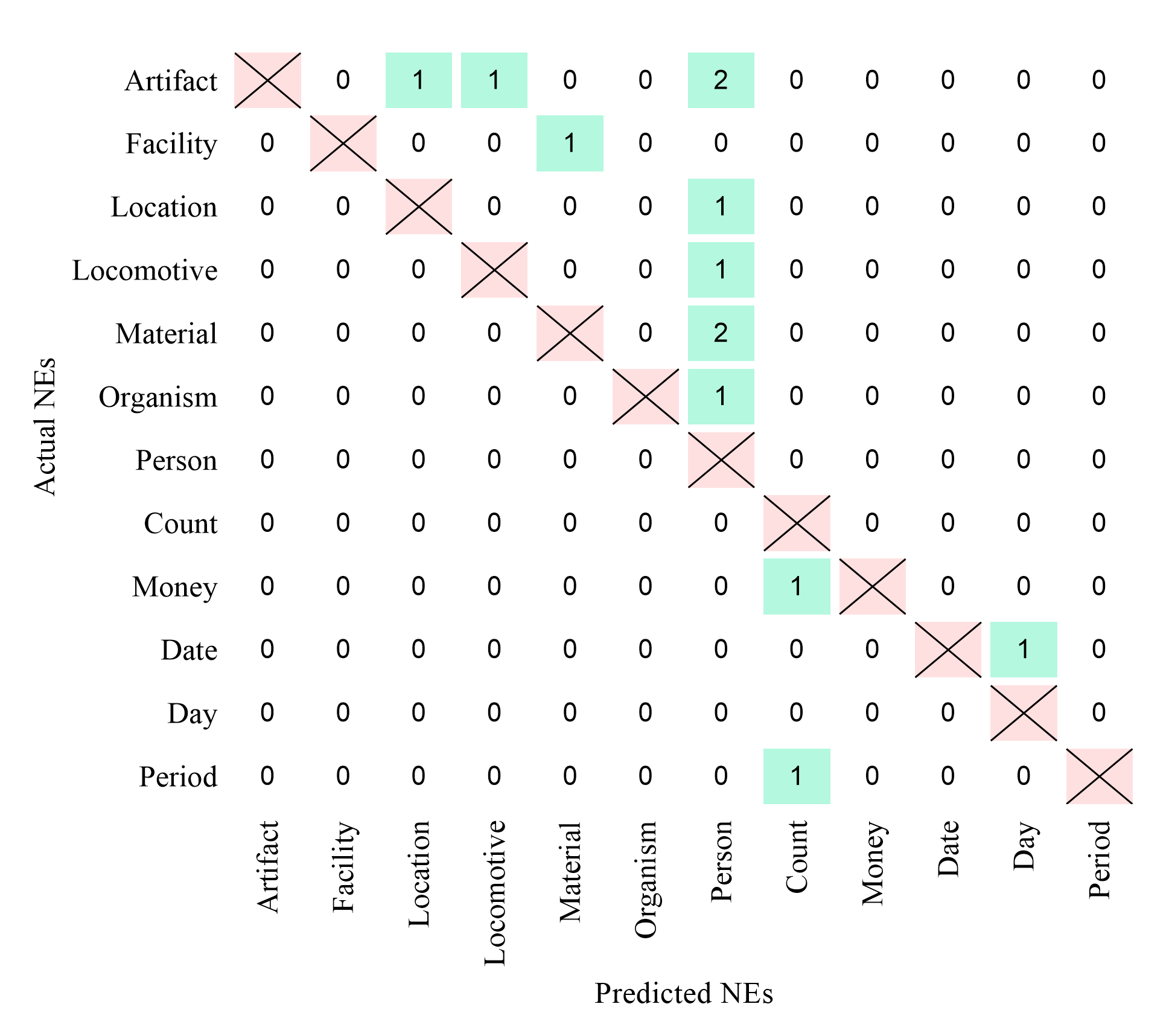}}
		\caption{Confusion matrix for Magahi for the LSTM-CNNs-CRF (above) and CRF (below) models; The \colorbox{pink!70}{$\times$} refers to correctly prediction}
		\label{mag_cm_nn_crf}
	\end{figure}

\end{document}